\documentclass[final]{cvpr}

\usepackage{times}
\usepackage{epsfig}
\usepackage{graphicx}
\usepackage{amsmath}
\usepackage{amssymb}

\usepackage{abstract}
\usepackage{amsmath}
\usepackage{graphicx}
\usepackage{lineno}
\usepackage{array}
\usepackage{longtable}
\usepackage[numbers]{natbib}
\usepackage[usenames,dvipsnames]{color}
\usepackage{amssymb}
\usepackage{textcomp}
\usepackage{url}
\usepackage{float}
\usepackage{multirow}
\usepackage{microtype}
\usepackage{xcolor}

\usepackage{times}
\usepackage{epsfig}
\usepackage{graphicx}
\usepackage{amsmath}
\usepackage{amssymb}

\usepackage{subfig}
\usepackage{gensymb} % For degree symbol
\usepackage{booktabs}
\usepackage{multirow, makecell}
\usepackage{caption}
\usepackage{nccmath}
\usepackage{graphicx}
\usepackage{enumerate}

\usepackage{gensymb}

\def\Put(#1,#2)#3{\leavevmode\makebox(0,0){\put(#1,#2){#3}}}

\usepackage[accsupp]{axessibility} 
\usepackage{amsthm}

\theoremstyle{definition}

\makeatletter
\renewcommand*{\@fnsymbol}[1]{\ensuremath{\ifcase#1\or  \dagger\or *\or \ddagger\or
   \mathsection\or \mathparagraph\or \|\or **\or \dagger\dagger
   \or \ddagger\ddagger \else\@ctrerr\fi}}
\makeatother
\usepackage[pagebackref=true,breaklinks=true,colorlinks,bookmarks=false]{hyperref}

\def\cvprPaperID{5758} % *** Enter the CVPR Paper ID here
\def\confYear{CVPR 2022}

\begin{document}

%%%%%%%%% TITLE
\title{PSMNet: Position-aware Stereo Merging Network for Room Layout Estimation}

\def\namespacing{15pt}
\def\emailspacing{3pt}
\author{Haiyan Wang$^{1,2}$\thanks{Work done while Haiyan Wang was an intern at Zillow.}\hspace{\namespacing}Will Hutchcroft$^{1*}$\hspace{\namespacing}Yuguang Li$^1$\thanks{Authors contributed equally.}\hspace{\namespacing}Zhiqiang Wan$^1$\hspace{\namespacing}Ivaylo Boyadzhiev$^1$\hspace{\namespacing}\\Yingli Tian$^2$\hspace{\namespacing}Sing Bing Kang$^1$\\
$^1$Zillow Group\hspace{30pt}$^2$The City College of New York\\
{\tt\small hwang005@citymail.cuny.edu,\hspace{\emailspacing}ytian@ccny.cuny.edu}\\
{\tt\small $\{$willhu,yuguangl,zhiqiangw,ivaylob,singbingk$\}$@zillowgroup.com}
}

\twocolumn[{%
\maketitle
\renewcommand\twocolumn[1][]{#1}%
   \vspace{-5mm} 
    \centering
    \includegraphics[width=1\linewidth]{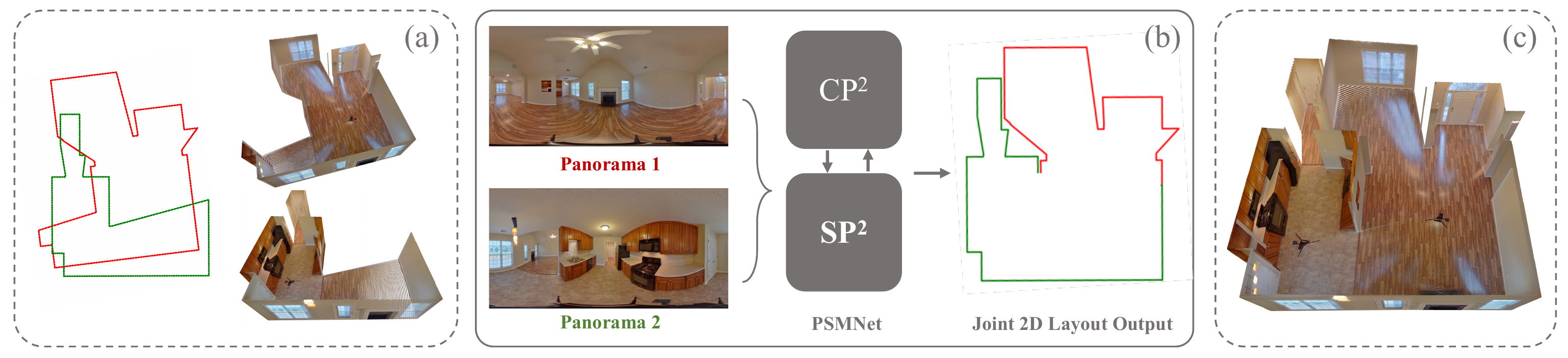} \captionsetup{type=figure}
    % \captionsetup{type=figure}
    \vspace{-5mm}
    \captionof{figure}{\footnotesize Estimating the complete layout of complex indoor spaces from a pair of 360$^\circ$ panoramas. 
    We use GT data for the sake of demonstration.
    Due to occlusion, a single panorama may view only a portion of the whole space. (a) shows the 2D and 3D room layout components, representing only portion of the whole space that is visible to each panorama. In practice, the input relative pose may be only approximately known; this is represented by the noisy alignment between the two partially visible components. Our proposed end-to-end PSMNet shown in (b) takes the two panoramas as input and jointly estimates the complete visible room layout in 2D, while refining a given noisy relative pose. (c) visualizes the estimated layout in 3D. (a) and (c) are used for visualization. The input and output of PSMNet are shown in (b).}
    \label{fig:motivation}
  \vspace{5mm} 
 }]
 
\pagestyle{empty}
\thispagestyle{empty}

\saythanks

%%%%%%%%% ABSTRACT
% \begin{abstract}
\centerline{\large\bf Abstract}
\vspace*{12pt}

{\it In this paper, we propose a new deep learning-based method for estimating room layout given a pair of 360$^\circ$ panoramas. Our system, called Position-aware Stereo Merging Network or PSMNet, is an end-to-end joint layout-pose estimator. PSMNet consists of a Stereo Pano Pose (SP$^2$) transformer and a novel Cross-Perspective Projection (CP$^2$) layer. The stereo-view SP$^2$ transformer is used to implicitly infer correspondences between views, and can handle noisy poses. The pose-aware CP$^2$ layer is designed to render features from the adjacent view to the anchor (reference) view, in order to perform view fusion and estimate the visible layout. Our experiments and analysis validate our method, which significantly outperforms the state-of-the-art layout estimators, especially for large and complex room spaces.}
\vspace{-3mm}
\vspace*{12pt}
% (stronger)
% \end{abstract}
%%%%%%%%% BODY TEXT

\section{Introduction}
\label{sec:intro}

Image-based room layout estimation is an important step to constructing models of home interiors for a variety of applications, such as virtual tours, path planning, floor plan generation, and home insights on square footage and architectural style. Much work has been done in room layout estimation, and current techniques perform well on simple Manhattan and Atlanta-world layouts. However, their performance degrade for large and complex rooms, e.g., those that have more than 10 corners.

It is not unusual (at least in North America) to find room layouts that are significantly more complex than cuboids or L-shapes. Examples include large open spaces with merged kitchen, dining room, and living room. The prevalence of complex rooms is evidenced by the statistics of real residential homes in ZInD~\cite{cruz2021zillow}. Figure~\ref{fig:motivation} illustrates the difficulty of layout estimation for a complex indoor space with many self-occlusions. Here, single image solutions would not be adequate due to occlusion. 
This is because no panorama is able to see the entire open space. Using both panoramas would, in principle, be able to better extract the layout. In addition, given that reliability is distance dependent (due to reductions in resolution at farther distances), such dependence is reduced with multiple views.

In this paper, we recover the room layout from two 360$^\circ$ panoramas. This has its challenges, because relative camera pose of the panorama pair needs to be  estimated jointly with layout. While techniques such as structure-from-motion exist, our goal is to generate  layouts of complex rooms that may lack features due to occlusions.

Wide baseline 2-view Structure from Motion (SfM) is still an open problem. In this work we assume that an input pose, potentially noisy, is provided. For example, this could be based on a rough user input~\cite{cruz2021zillow} or matching corresponding semantic elements with noisy predictions~\cite{Shabani_2021_ICCV}.

Our solution is a joint pose-layout deep architecture to predict 2D room layout and refine a noisy 3 DOF relative camera pose in an end-to-end manner. Our system, called Position-aware Stereo Merging Network (PSMNet), consists of a transformer-based Stereo Pose Estimation (SP$^2$) network  and a new pose-aware Cross-Perspective Projection (CP$^2$) module. CP$^2$ generates the final layout with the help of an attention-based merging model (inspired by SEBlock~\cite{hu2018squeeze}) that weights regions based on certainty. The pose and layout modules share the same encoder for efficiency, and are trained end-to-end.

In our work, we make the same assumptions as ZInD~\cite{cruz2021zillow}: input panoramas are both captured upright at approximate same height, and layouts are based on Atlanta world (horizontal floor and ceiling, and vertical walls). The ceiling height is used for visualization.

The contributions of our work are:
\vspace{-1.2mm}
\begin{enumerate}[(i)]
\item First end-to-end joint layout-pose deep architecture (to our knowledge) for large and complex room layout estimation from a pair of panoramas. 
\vspace{-2mm}

\item New Cross-Perspective Projection (CP$^2$) module with attention-based merging for layout generation.

\vspace{-2mm}
\item An integrated transformer-based relative Stereo Pano Pose (SP$^2$) network to refine noisy input pose.
\vspace{-2mm}

\item State-of-the-art performance on a challenging, stereo panoramas dataset, sampled from ZInD~\cite{cruz2021zillow}.
% \vspace{-3mm}

\end{enumerate}

%------------------------------------------------------------------------
\section{Related Work}
\label{sec:related}

\begin{figure*}[ht]
\begin{center}
\includegraphics[width=1\linewidth]{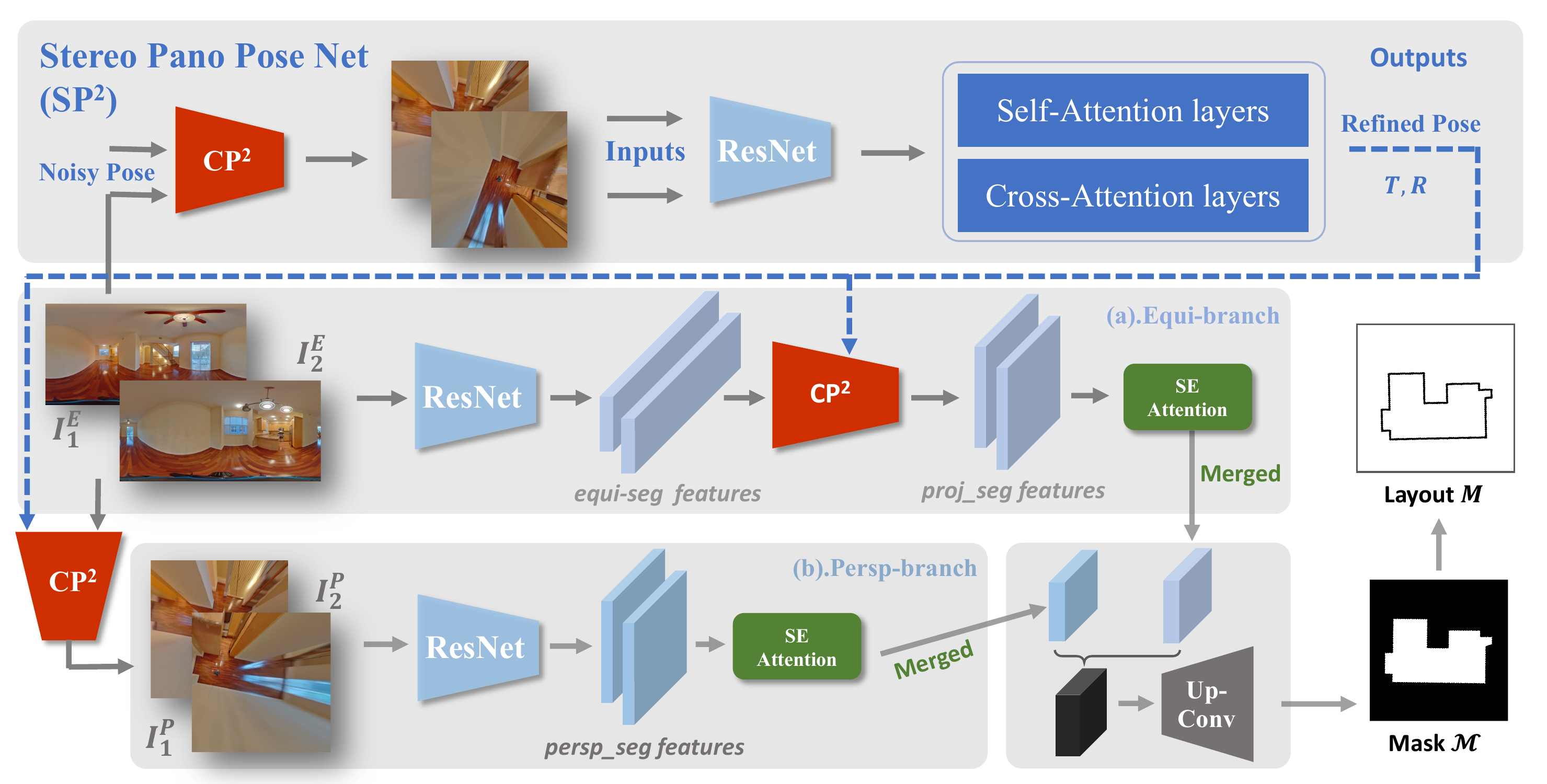}
\vspace{-5pt}
\caption{ Our proposed PSMNet architecture.
Its input is a pair of equirectangular panoramas with noisy relative pose, from which perspective projections are generated by the Cross-Perspective Projection layer (CP$^2$) as additional inputs to generate the room layout. Stereo Pano Pose (SP$^2$) Net  is trained to refine the relative pose.
The output of PSMNet is mask $\cal M$, which is then post-processed to generate the layout polygon $M$.
}
\vspace{-5mm}
\label{fig:pipeline}
\end{center}
\end{figure*}

In this section, we review approaches relevant to our work. They are organized based on the following attributes for room layout estimation: (1) partial versus complete layouts, (2) single-view, and (3) multi-view 360$^\circ$ panoramas. More extensive surveys can be found in \cite{Mathew2020ReviewOR} and \cite{PMGFPG20a}.

\subsection{Partial vs Complete Rooms Layouts}

Much work has been done on generating partial room layout from a single perspective image \cite{Chao2013LayoutEO, Delage2006ADB, Hedau2009RecoveringTS, Lee2017RoomNetER, Lee2009GeometricRF,  Schwing2012EfficientSP,   Zheng2020StructuralDM, zou2018layoutnet}. Early, geometric-based approaches analyze lines and vanishing points \cite{Flint2010ADP, Hedau2009RecoveringTS}. With the introduction of large-scale datasets \cite{cruz2021zillow, Hedau2009RecoveringTS, zhang2015large, Zheng2020Structured3DAL}, most of the recent work is learning-based~\cite{Delage2006ADB, Lee2017RoomNetER, Zheng2020StructuralDM}.

Extending \cite{Flint2010ADP} to multiple perspective views, \cite{Flint2011ManhattanSU} proposed a hybrid approach where low-level cues (extracted from structure-from-motion) are combined into a learnable Bayesian framework to build a multi-view consistent partial room layout. Extending further the input requirements, by using a small number of overlapping perspective RGB-D images, \cite{Lin2019FloorplanJigsawJE} proposed a geometric-based approach to fuse multiple, partial pieces into a complete room layout. 

\subsection{Single-view 360$^{\circ}$ Layout Estimation }

PanoContext \cite{Zhang2014PanoContextAW} was one of the first to study the effect of FoV for room layout estimation. Similar to other early work \cite{Yang2016Efficient3R}, they first convert a single panorama into overlapping perspective images to estimate per-pixel normals, by combining \cite{Lee2009GeometricRF} and \cite{Hedau2009RecoveringTS}, which is later used to evaluate room layout hypothesis.
LayoutNet \cite{zou2018layoutnet} demonstrated the benefits of operating directly on the equirectangular panorama. They use an encoder-decoder CNN, similar to RoomNet \cite{Lee2017RoomNetER}, to estimate the corner and boundary probabilities for cuboid layout estimation. DuLa-Net \cite{yang2019dula} jointly exploited the equirectangular panorama and its perspective ceiling-view in an end-to-end differentiable network, using a novel equirectangular-to-perspective (E2P) feature fusion step.

HorizonNet \cite{sun2019horizonnet} is a seminal approach that generates a compact 1D representation where each image column of the equirectangular panorama encodes the floor-wall, ceiling-wall, and wall-wall boundaries. A bidirectional RNN is used to learn short and long-term dependencies across the panorama. Many subsequent approaches \cite{Rao2021OmniLayoutRL, sun2021hohonet, Tran2021SSLayout360SI, wang2021led2} adopted HorizonNet as their back-bone architecture. AtlantaNet \cite{pintore2020atlantanet} proposed floor and ceiling-view projections to combine the benefits of DuLa-Net and HorizonNet. They handle the more complex Atlanta-world cases \cite{Schindler2004AtlantaWA}. Despite the increased 360$^\circ$ FoV, monocular layout estimation techniques are less effective for large open spaces or complex rooms with self-occlusion.

\subsection{Multi-view 360$^{\circ}$ Layout Estimation }

Most techniques on multi-view 360$^{\circ}$ layout estimation are focused on full floor-plan reconstruction from a sparse set of overlapping RGB panoramas \cite{Cabral2014PiecewisePA, Pintore20183DFP} or a dense sequence of RGB-D scans \cite{Chen2019FloorSPIC, Fang2021FloorplanGF, Fang2021StructureawareIS, Liu2018FloorNetAU}. The pure RGB approaches \cite{Cabral2014PiecewisePA, Pintore20183DFP} typically start with SfM \cite{Kangni2007OrientationAP} to determine relative panorama poses. However, this step tends to fail for sparsely captured panoramas with wide baselines, as demonstrated by \cite{cruz2021zillow, Shabani_2021_ICCV}. Assuming all images can be localized,
a common approach is to first segment each input image into floor, wall and ceilings regions (akin to \cite{Yang2016Efficient3R}), in combination with multi-view cues and constraints \cite{Cabral2014PiecewisePA, Pintore20183DFP, Pintore2018Recovering3E}. The multi-view segmentation maps are then projected, as 2D layouts, and fused together into a final floor-plan boundary \cite{Cabral2014PiecewisePA}, or per-room layout boundaries \cite{PGJG19, Pintore2018Recovering3E}.

Recently, \cite{muti-viewpano} proposed a multi-view layout reconstruction for large indoor spaces, starting from multiple panorama images with known poses. Using ideas from DuLa-Net and HorizonNet, they first use \cite{sun2019horizonnet} to obtain single-view layout predictions, which are then converted into ceiling-view segmentation masks. Their key idea is to train a DNN to generate multiple ceiling-wall (boxification) line proposals from each view. Those are then fused using a graph-cut optimization to obtain a single multi-view consistent 2D layout. Their method can handle more than 2 views. However, they highly rely on the quality of the pre-computed single-view layouts as well as the given camera poses.

In contrast, we focus on 2-view layout reconstruction, with approximate poses. We propose an end-to-end fully differentiable DNN to jointly estimate multi-view consistent layout while refining pose.

%------------------------------------------------------------------------

\section{Framework}
\label{sec:overview}
In this section, we describe PSMNet (Figure~\ref{fig:pipeline}) and the loss function we optimize.
PSMNet includes the Cross-Perspective Projection layer (CP$^2$) and Stereo Pano Pose Net (SP$^2$), both of which are described in subsequent sections.

\subsection{Architecture Design}\label{sec:architecture}

PSMNet adopts a backbone similar to DuLa-Net \cite{yang2019dula} or AtlantaNet \cite{pintore2020atlantanet} with a dual-segmentation structure. The inputs to PSMNet are two 360$^{\circ}$ panoramas $I_1^E, I_2^E$ with camera viewpoints $V_1, V_2$, respectively. Without loss of generality, let the first be the anchor view, with the other (secondary view) the target for pose estimation and feature fusion. Each image $I_n^{E} (n=1,2)$ is processed in equirectangular space and perspectively projected to the anchor view. The latter operation, which we call Cross-Perspective Projection (CP$^2$) layer , uses the relative pose estimated by the Stereo Pano Pose (SP$^2$) Network given the two panoramas. 
%We use our 

Given a potentially noisy input pose, SP$^2$Net uses a transformer-based attention mechanism to refine the relative positions between stereo-view panoramas. We also extract two sets of segmentation features, namely \textit{equi-seg features} and \textit{persp-seg features}, with each set being the result of concatenating two views. The \textit{equi-seg features} are further rendered to the anchor view in the perspective space in the same way as was done for the panoramas, resulting in \textit{proj-seg features}. Note that \textit{persp-seg features} and \textit{proj-seg features} are camera aligned and thus can readily be merged. 

Prior to segmentation feature merging, we apply an attention model inspired by \cite{hu2018squeeze} to extract an implicit confidence representation. As the concatenated segmentation features are generated from two separate camera views, each view's feature vector encodes content from differing regions of the room. As a result, the contributions of these features on the final merged feature are expected to be non-uniform (e.g., due to depth and texture variation). An SE-Attention model is used to estimate these contribution weights. The refined \textit{persp-seg features} and \textit{proj-seg features} are concatenated and fed to a set of Up-Conv layers. The output of the last Up-Conv layer is binary mask $\cal M$ of the 2D room layout.

The final polygon layout is generated by our proposed Mostly Manhattan algorithm as follows. We first extract a dense contour from $\cal M$, which is then fit with line segments using the Douglas-Peucker algorithm~\cite{saalfeld1999topologically}. While the majority of published work imposed a strong Manhattan constraint on the post-processed layout, we allow some walls to be non-Manhattan when a candidate wall is greater than a threshold $\gamma$ away from one of the coordinate axes; more details are found in the supplementary material. As we do not estimate the ceiling height, the 3D layout can be extracted by extruding the layout with the ground truth ceiling height.

\subsection{Loss Function Design}

PM$^2$Net is jointly trained end-to-end on the pose and layout estimation tasks. The pose estimation is formulated as a regression problem; we compute the $\ell_1$ loss between the ground truth and predicted pose parameters. We denote the rotation loss by $L_{\rm p}^{(R)}$ and translation loss by $L_{\rm p}^{(T)}$. The pose loss is given by
\begin{equation}\label{eq:loss_p}
L_{\rm p} = \mu L_{\rm p}^{(R)} + (1-\mu) L_{\rm p}^{(T)}.
\end{equation}

Layout estimation is cast as a segmentation process where we compute the cross-entropy losses between the predicted room shape mask and the ground truth in both equirectangular and perspective spaces, denoted by $L_{l}^{(E)}$ and $L_{l}^{(P)}$, respectively. The layout loss is
\begin{equation}\label{eq:loss_s}
L_{l} = L_{l}^{(E)} + L_{ l}^{(P)}.
\end{equation}

The total loss for end-to-end training is
\begin{equation}\label{eq:loss_tot}
L_{\rm tot}=(1-\lambda)L_{\rm p}+\lambda L_{l}.
\end{equation}
Note that in Eqs. \eqref{eq:loss_p} and \eqref{eq:loss_tot}, the hyper-parameters $\mu,\lambda\in [0,1]$.
For our experiments, both $\mu$ and $\lambda$ are set as 0.5.

%------------------------------------------------------------------------

\section{Cross-Perspective Projection (CP$^2$) Layer} \label{sec:mp-layer}

PSMNet augments the equirectangular panoramas $I_n^{E} (n=1,2)$ with aligned perspective-projected top-down views as additional signals. These top-down views are generated by the CP$^2$ layer, using the first (anchor) viewpoint as reference; let these views be $I_n^{P} (n=1,2)$. The two panoramas are assumed to be axis aligned vertically \cite{Zhang2014PanoContextAW}. We use normalized texture coordinates 
$(u_{n}^{E}, v_{n}^{E}, n=1,2)$ ranging from 0 to 1 to represent position in $ I_1^{E} $ and $ I_2^{E} $.

The 3 DOF pose of the secondary relative to the anchor is $\{ \Delta x, \Delta y, \Delta \theta \}$, indicating relative 2D position shift and horizontal angular difference. These are refined by SP$^2$Net as described in Section~\ref{sec:pose}.

Let the focal lengths of $I_n^{P} (n=1,2)$ be $f_n^{P} (n=1,2)$. The panorama south pole can be found at the origin of $I_{1}^{P}$. $I_{2}^{P}$ is projected from $ I_2^{E}$ to be in the same reference coordinate system as $I_{1}^{P}$. For a fixed field of view $FoV_1$ in the anchor view, $f_1^{P}$, $FoV_2$ (field of view in the secondary view), and $f_2^{P}$ can be found as follows:
\begin{equation}
f_1^{P}=0.5 \lambda \cot(0.5 FoV_1),
\end{equation}
\begin{equation}\label{eq:cam_h_to_focal_length}
FoV_2 = 2 \tan^{-1}(\frac{H_{1}^{E}}{H_{2}^{E}} \tan(0.5 * FoV_1)),
\end{equation}
\begin{equation}\label{eq:fov_to_focal_length}
f_2^{P}=0.5 \lambda \cot(0.5 FoV_2),
\end{equation}
% \yuguang{Equation (4) and (6) are repeating themselves. Is there a way to merge them into 1}
where $H_n^{E} (n=1,2)$ are the camera heights and
$\lambda$ is the width (in pixels) of $I_n^{P} (n=1,2)$. In our work, we assume $H_1^{E} = H_2^{E}$, which results in $FoV_1 = FoV_2$ and $f_1^{P} = f_2^{P}$.

\begin{figure}[t]
\begin{center}
\includegraphics[width=1\linewidth]{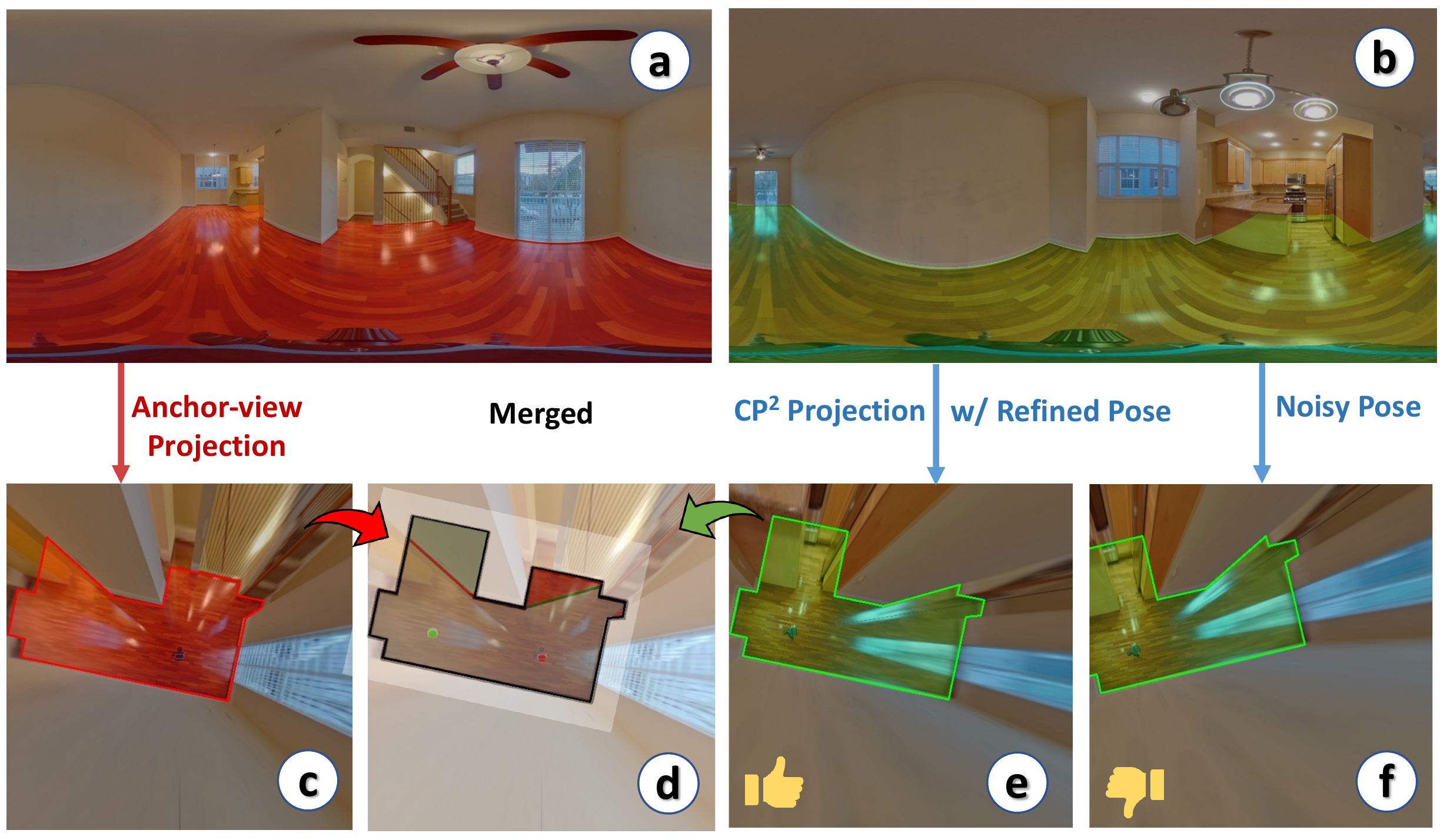}
% \vspace{-5pt}
\caption{ \small Illustration of our proposed CP$^2$. We plot visible floor area on the panorama and perspective projected images. (c) is the anchor-view projection from (a). (d) displays the estimated room layout from both views with jointly refined camera pose. (e) and (f) compare the projected adjacent panorama to the anchor view with and without the pose refinement, respectively.}
\vspace{-5mm}
\label{fig:cp2}
\end{center}
\end{figure}

Given translation $t_n = \{ x_n, y_n \} $ and rotation $\theta_n$, $n=1,2$, our CP$^2$ layer projects the panorama coordinates as follows:

\begin{equation}\label{eq:u2_from_texture}
u_n^{E} = \frac{atan2(p_x^{n} - x_n, p_y^{n} - y_n) -  \theta_n}{2\pi },
\end{equation}
\begin{equation}\label{eq:v2_from_texture}
v_n^{E} =1-\frac{atan2(\left \| p_x^{n} - x_n, p_y^{n} - y_n \right \|_{2}, f_n^P)}{\pi },
\end{equation}
where $(p_{x}^{n}, p_{y}^{n}, n=1,2)$ is a point in the \textit{joint floor coordinate} system.
For the anchor view, $x_1 = y_1 = \theta_1 = 0$. For the secondary view, $x_2 = \Delta x, y_2 = \Delta y, \theta_2 = \Delta \theta$. The effect of CP$^2$ is illustrated in the Figure \ref{fig:cp2}.

%-----------------------------
\section{Stereo Panorama Pose (SP$^2$) Network} \label{sec:pose}

We assume that the relative pose between the two input panoramas is only approximately known; in practice, any pose estimate will be subject to noise. The SP$^2$Net component of PSMNet is responsible for refining the initial pose estimate. More specifically, the goal of SP$^2$Net is to predict the pose refinement parameters $ \Delta t = t_{gt} - t_c  $ and $ \Delta \theta = \theta_{gt} - \theta_c  $, where $ \Delta t = \{ \Delta x, \Delta y \} $, ($t_c$, $\theta_c$) is the input noisy pose and ($t_{gt}$, $\theta_{gt}$) the ground truth pose.

First, the anchor panorama $ I_1^{E} $ is projected to perspective view $ I_1^{P} $ by the CP$^2$ layer, with the pose parameters all set to 0. The second input panorama $ I_2^{E} $ is projected to perspective view $ I_2^{P} $ by the same CP$^2$ layer using the input noisy pose ($t_c$, $\theta_c$).
The shared backbone ResNet-18 is then used to extract multi-scale features from $ I_1^{P} $ and $ I_2^{P} $. The extracted features are denoted as $ F_1^{P} $ and $ F_2^{P} $.

$ F_1^{P} $ and $ F_2^{P} $ are added with positional encodings, and each feature map is then flattened to a 1-D vector.  
The encoded features are passed through a transformer to extract position and context dependent local features. The transformer (inspired by \cite{sun2021loftr}) consists of self-attention and cross-attention layers. The output features from the transformer are denoted as $ T_1^{P} $ and $ T_2^{P} $.

Finally, $ T_1^{P} $ and $ T_2^{P} $ are concatenated along the channel dimension. The concatenated features $ C^{P} $ are fed into three convolutional layers to extract features that contain the relative pose information between the two input panoramas. The extracted features are flattened, and a fully connected layer is used to predict $ \Delta t $ and $ \Delta \theta $.
% \vspace{-2mm}

%------------------------------------------------------------------------
\section{Experiments} \label{sec:experiments}

In this section, we report results of our approach on Zillow Indoor Dataset (ZInD) \cite{cruz2021zillow}. Given the variability of relative location of panoramas as well as the complexity of the room layout, we use \emph{spatial overlap} and \emph{co-visibility} (which we define shortly) to stratify our results.

%\subsection{Datasets}

Most room layout datasets such as PanoContext \cite{Zhang2014PanoContextAW}, Stanford 2D-3D \cite{zou2018layoutnet}, and Realtor360 \cite{yang2019dula} only feature a single panorama per room, making them not suitable for our work. We chose Zillow Indoor Dataset (ZInD)~\cite{cruz2021zillow} for evaluation because it is the only large-scale, public dataset that has the multi-view panorama configuration for layout estimation, it is based on many real residential homes across many cities, and its rooms have significant geometric diversity (Manhattan and non-Manhattan, and significant spread in room size and number of room corners). We derive a stereo-view dataset from ZInD for our experiments. In total, there are 107,916, 13,189, and 12,348 pano pair instances in our train, test, and val splits, from 40,336, 5,138, and 4,993 unique panoramas, respectively.

% \subsubsection{Ablation of camera distance}

\begin{figure}[t]
\begin{center}
\includegraphics[width=1\linewidth]{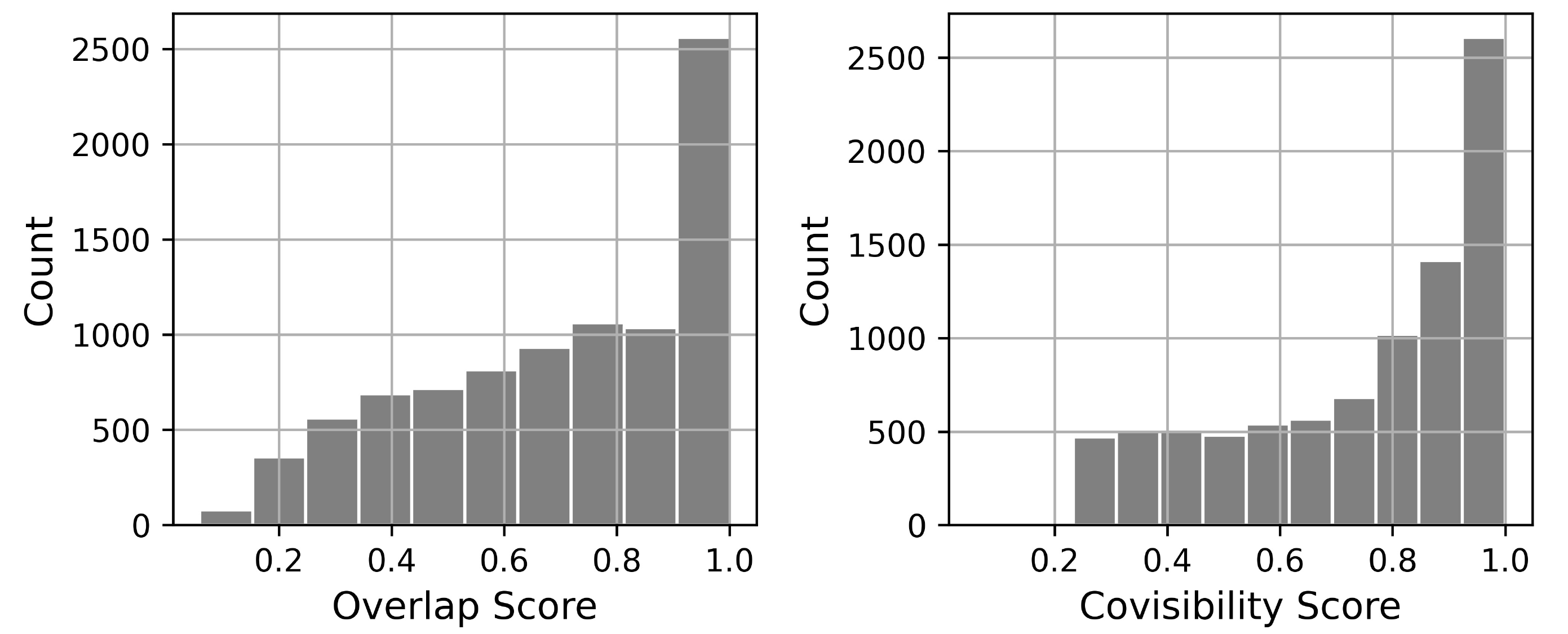}
% \vspace{-5pt}
\caption{ \small Data distribution of the spatial overlap and co-visibility scores. 
}
\vspace{-8mm}
\label{fig:data}
\end{center}
\end{figure}

\subsection{Benchmarks and Evaluation Metrics}
\label{sec:benchmarks}

We compare our PSMNet with baselines built upon recent state-of-the-art layout estimation methods: HorizonNet~\cite{sun2019horizonnet}, DulaNet~\cite{yang2019dula}, LED$^2$Net~\cite{wang2021led2} and HoHoNet~\cite{sun2021hohonet}. Since these methods only address single view layout estimation, we first derive the estimated result for each view and then perform a simple shape union across stereo views to get the merged result. Note that room layout recovery is based on what is {\em visible}; an occlusion edge shows up as a ``wall".

In deriving single-view results, we found that using fully-Manhattan post-processing decreases the baseline performance. This is because ZInD contains many partially visible layouts, which introduces non-Manhattan occlusion ``walls". 

To increase the baseline performance, we instead apply post-processing which preserves non-Manhattan structure. For HorizonNet and HoHoNet, we sample the predicted contour at corner lines of sight to get the final layout polygon. For the segmentation based methods, we apply AtlantaNet's post-processing, which also preserves non-Manhattan walls. For all methods, we apply our Mostly Manhattan post-processing (Section~\ref{sec:architecture}) to the merged shape union to get the final room layout. The quality of the stereo-view layout estimate is evaluated using \emph{2D IoU}.
We also use $\delta_i$~\cite{yang2019dula}, which measures accuracy in panorama pixel space.

\emph{Spatial overlap} and  \emph{co-visibility} are employed in our experiments as measures of difficulty and to stratify results. For a panorama pair, \emph{spatial overlap} measures the IoU between the ground truth single-view visible layout polygons. The higher the score, the more visible floor area the two panoramas have in common. As shown in Figure~\ref{fig:data}, we split the dataset into 
\emph{Overlap-High} ($>$ 0.9), \emph{Overlap-Medium} (0.5 - 0.9) and \emph{Overlap-Low} ($<$ 0.5). In the test set, each split has 3,769, 5,644, and 3,776 data instances, respectively. 

Since our method incorporates both the perspective and equirectangular projections, we additionally stratify by \emph{co-visibility}~\cite{cruz2021zillow}, which measures visual overlap ($\in [0,1]$) between two panoramas. Examples are split into \emph{Covis-High} ($>$ 0.9), \emph{Covis-Medium} (0.5 - 0.9), and  \emph{Covis-Low} ($<$ 0.5).

\begin{table*}[htbp]
  \centering 
  \caption{ Quantitative evaluation stratified by spatial overlap at different levels of room complexity. Note that "Overlap-Low" indicates higher room complexity with more occlusions.}
  \scalebox{0.97}{
    \begin{tabular}{c||c|cc|cc|cc|cc}
    \hline
    \multirowcell{2}{Pose} & \multirowcell{2}{Methods} & \multicolumn{2}{c|}{Overall} & \multicolumn{2}{c|}{Overlap-High} & \multicolumn{2}{c|}{Overlap-Medium} & \multicolumn{2}{c}{Overlap-Low} \\
\cline{3-10}          &       & 2D IoU ($\%$) &  $\delta_i$  & 2D IoU ($\%$)&  $\delta_i$  & 2D IoU ($\%$) &  $\delta_i$   & 2D IoU ($\%$)&  $\delta_i$   \\\hline\hline
    \multicolumn{1}{c||}{\multirowcell{5}{w/ \\ GT}} 
    & DulaNet~\cite{yang2019dula} &  64.03     & 0.8043     & 65.21    & 0.8185     & 62.14    & 0.8031    & 60.27   & 0.7980 \\
    & HorizonNet~\cite{sun2019horizonnet}  & 73.35    & 0.8663 & 82.08 & 0.8801 & 71.47 & 0.8678 & 69.20 & 0.8535 \\
          & HoHoNet~\cite{sun2021hohonet} & 74.25 & 0.8649 &  82.55   & 0.8816     & 72.43   & 0.8672     & 70.35    & 0.8486 \\
          & LED$^2$Net~\cite{wang2021led2} & 76.39 & 0.9056 & 83.68 & 0.9243 & 73.73 & 0.8736 & 72.08 & 0.8697 \\
          & PSMNet (Ours) & \textbf{81.01} & \textbf{0.9238} & \textbf{85.71} & \textbf{0.9349} & \textbf{80.13} & \textbf{0.9253} & \textbf{76.93} & \textbf{0.9074} \\\hline
    \multicolumn{1}{c||}{\multirowcell{5}{w/o \\ GT}} 
          & DulaNet~\cite{yang2019dula} &  59.30     & 0.7828     & 62.06     & 0.7699    & 57.97    & 0.7855    & 51.21    & 0.7936 \\
          & HorizonNet~\cite{sun2019horizonnet} & 62.79 & 0.8354 & 70.98 & 0.8437 & 61.51 & 0.8355 & 58.24 & 0.8288 \\
          & HoHoNet~\cite{sun2021hohonet} & 63.31 & 0.8324 & 70.59    & 0.8390  & 62.03     & 0.8339    & 59.47   & 0.8253\\
          & LED$^2$Net~\cite{wang2021led2} & 65.81 & 0.8566 & 71.06 & 0.8493 & 64.81 & 0.8574 & 63.14 & 0.8611 \\
          & PSMNet (Ours) & \textbf{75.77} & \textbf{0.9217} & \textbf{84.80} & \textbf{0.9371} & \textbf{74.73} & \textbf{0.9210} & \textbf{66.73} & \textbf{0.9040} \\
    \hline
    \end{tabular}%
    }
  \label{tab:covis}%
\end{table*}%

\subsection{Implementation Details}

PSMNet is implemented in PyTorch and trained with the Adam \cite{kingma2014adam} optimizer on a single GPU for 200 epochs. We set the learning rate as 0.0001 and the batch size as 6. The backbone feature extraction network is ResNet18~\cite{He_2016_CVPR}. The Manhattan threshold $\gamma$ in Section~\ref{sec:architecture} is set to 10. 

The joint layout-pose network is trained with two configurations (one with ground truth pose and the other with noisy pose augmentation). For pose noise, in the training stage, we perturb the ground truth pose with noise sampled from a uniform distribution. We perform further data augmentation by randomly switching which panorama is selected as the anchor view. For all of the single-view baseline models, we re-train the network on single-view examples from the ZInD stereo dataset, with 200 epochs for fair comparison. After getting the single-view layout estimation results, we apply both the same ground truth pose and noisy pose to perform the merging process.

To assess our model's tolerance to noise, we synthetically generate noisy pose estimates by sampling. In practice, any pose estimator may be used, such as LayoutLoc~\cite{cruz2021zillow}. 2-view wide baseline SfM remains a challenging open problem.

\subsection{Quantitative Evaluation} \label{sec:Quantitative}

In our evaluation of PSMNet on the ZInD stereo-view dataset, we apply the same ground truth (GT) and noisy poses as inputs for other baseline methods, as well as PSMNet, for an apples-to-apples comparison. Quantitative results reported in Table~\ref{tab:covis} show that PSMNet consistently outperforms the baseline methods with significant improvements especially for complex rooms.
\vspace{-3mm}

\paragraph{Performance with GT pose.} %\label{sec:LayoutGT}
With known GT pose, the influence of pose refinement removed. As shown in the upper half of Table~\ref{tab:covis}, with GT pose as input, PSMNet shows an overall improvement over the LED$^2$Net baseline of 4.62$\%$ for 2D IoU and 0.02 for $\delta_i$. Overlap High demonstrates the most competitive baseline performance. With high visual overlap, the benefit of an additional view is reduced. Nevertheless, we improve upon all baselines. The advantage of PSMNet further increases as visual overlap is reduced (Overlap-Medium and Overlap-Low).
\vspace{-3mm}

\paragraph{Performance with noisy pose.} %\label{sec:LayoutLayoutLoc}
To assess noise tolerance, we sample pose noise from uniform distributions $\mathbf{U}(0, 40^\circ)$ and $\mathbf{U}(0, 1m)$, for rotation and translation, respectively. As shown in the lower part of Table~\ref{tab:covis}, when the input pose is noisy, our joint layout-pose estimation pipeline surpasses the baseline performance by a large margin. 
For Overlap-High, even without GT pose, PSMNet performs better than most baselines {\em with GT pose}. This is compelling illustration of the benefit of SP$^2$.
Additional results for all methods stratified by \emph{co-visibility} are shown in the supplementary. They are similarly positive for PSMNet.
\vspace{-3mm}

\begin{figure}[t]
\begin{center}
\includegraphics[width=1\linewidth]{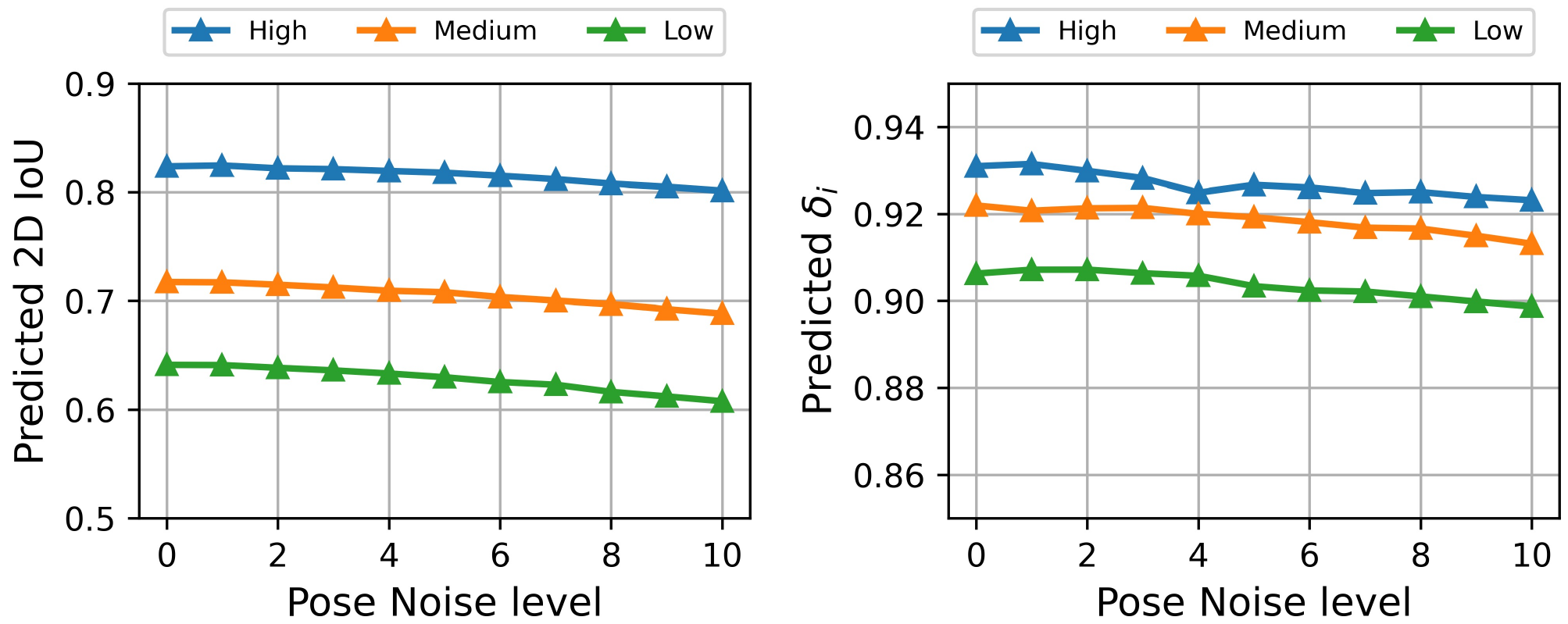}

\caption{  Illustration of the robustness of our proposed PSMNet under various level of pose noise for Overlap-High, Overlap-Medium, and Overlap-Low groups.}
\vspace{-5mm}
\label{fig:coarse}
\end{center}
\end{figure}

\begin{figure*}[ht]
\begin{center}
\includegraphics[width=\textwidth]{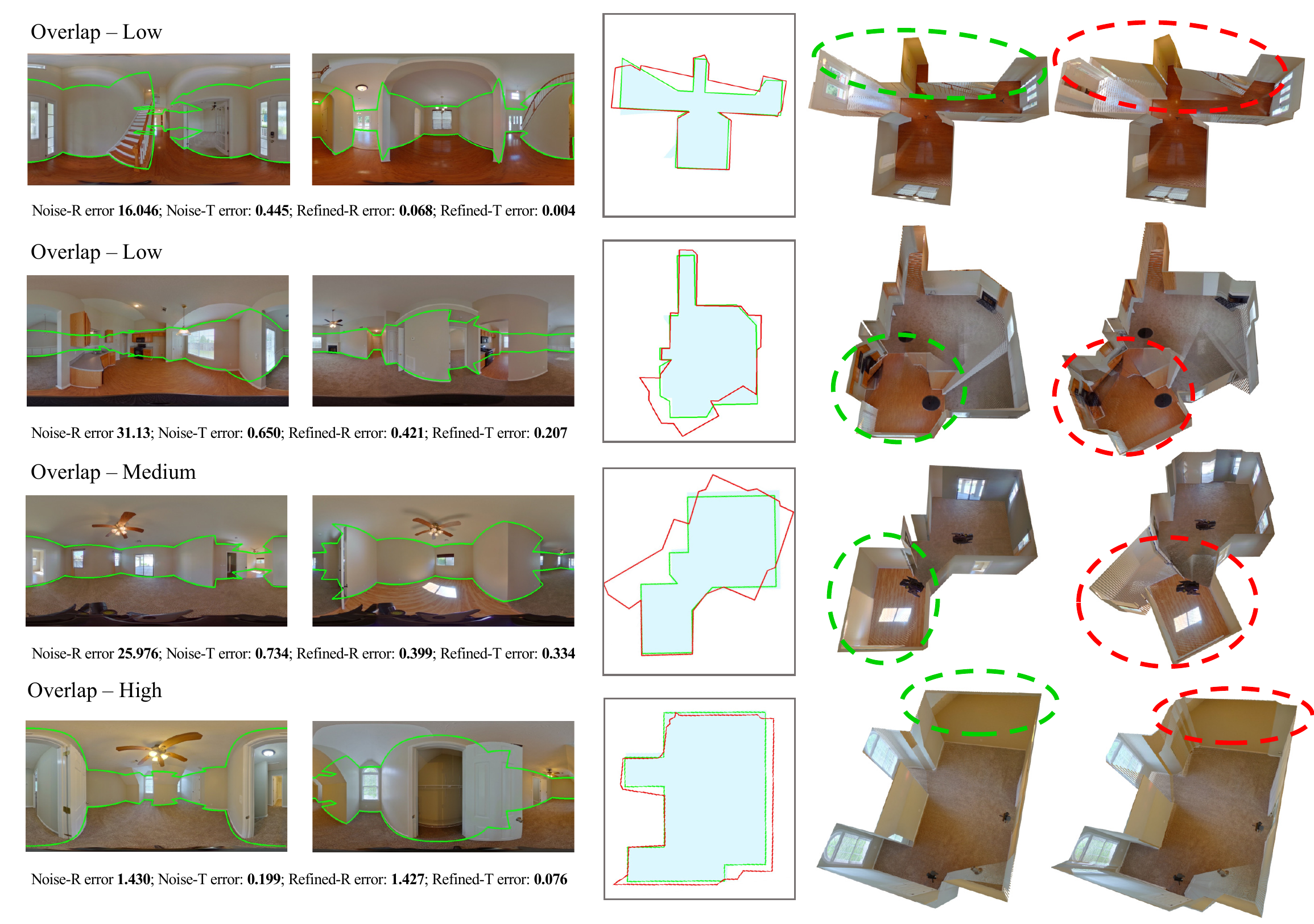} 
% \vspace{-5pt}
\caption{ Position-aware layout estimation results on the ZInD dataset with noisy pose. The first two columns show the estimated visible room layout on each panorama in {\color{green}green}. In the middle top-down plot we stack the predicted room shape in {\color{green}green} over the ground truth mask in {\color{cyan}cyan}, while the LED$^2$Net baseline result is shown in {\color{red}red}. The fourth column demonstrates the 3D layouts of our PSMNet and the last column is the results of LED$^2$Net.}
\vspace{-5mm}
\label{fig:results}
\end{center}
\end{figure*}

\paragraph{Effect of pose noise.}
Our simulated coarse poses are generated by adding $\pm \theta_{err}^{\circ}$ rotational errors as well as translation errors of fixed magnitudes in a random direction. Figure~\ref{fig:coarse} shows our system performance with pose noises ranging from $0 \rightarrow (0m, 0^\circ)$ to $10 \rightarrow (1m, 40^\circ)$, with increments of $(0.1m, 4^\circ)$. We compute both 2D IoU and $\delta_i$ on the stratified ZInD dataset. PSMNet demonstrates robust performance over a range of input pose noises, with higher than 80$\%$ IoU on \emph{Overlap-High} and more than 60$\%$ IoU on \emph{Overlap-Low}, even when the pose noise increases significantly to $(1m, 40^\circ)$. The $\delta_i$ plot shows a similar trend.

\subsection{Qualitative Evaluation}
In Figure~\ref{fig:results}, we show sample estimated layouts for PSMNet compared with the LED$^2$Net baseline. In the first two columns, the accuracy of both our layout and pose refinement is demonstrated by the alignment with the true boundaries in the equirectangular images, for both the anchor and adjacent view alike. Note that while they are captured in segmentation, we do not post-process additional ``internal" polygons for islands, pillars, or separating walls that arise in large spaces. This can be seen by the absence of partial boundary in row 2. We further compare {\color{green}PSMNet}, the {\color{red}LED$^2$Net} baseline, and {\color{cyan}GT}, displayed in top-down projection, in column 3. Columns 4 and 5 show the 3D layouts for PSMNet and LED$^2$Net, extruded by GT ceiling height for display. The benefits of SP$^2$, and by contrast the consequences of pose noise, are striking. We highlight significant differences; the benefit of end-to-end learning of pose and layout is most noticeable when comparing the cohesive PSMNet layouts, with the poorly merged regions of the LED$^2$Net baseline. 
% \vspace{-5pt}

\subsection{Ablation Study on Network Components}

\begin{table}[t]
  \centering
  \caption{Evaluation of different variants of PSMNet.}
    \begin{tabular}{ccc||cc}
    \hline
    CP$^2$  & SP$^2$ & SE Attention & 2D IoU ($\%$) & $\delta_i$ \\
    \hline
    \hline
    $\times$    & $\times$     & $\times$     &  \underline{60.41}     & \underline{0.8031} \\
    \checkmark     & $\times$      & $\times$     &  66.39     & 0.8701 \\
    \checkmark     & \checkmark     & $\times$     &  70.92     & 0.8883 \\
    \checkmark     & $\times$     & \checkmark    &   69.14    & 0.8723 \\
   \checkmark     & \checkmark    & \checkmark    &   \textbf{72.38}    & \textbf{0.9003} \\
    \hline
    \end{tabular}%
    \vspace{-2mm}
  \label{tab:varient}%
\end{table}%

We conduct experiments to investigate the affect of individual components in our PSMNet architecture. Specifically, we consider the following variants: 
\vspace{-2mm}

\begin{enumerate}[(i)]

\item Remove the proposed CP$^2$,  SP$^2$, and SE Attention layers. Instead of cross-perspective rendering, we do a direct perspective rendering for the second view of input panorama $I_2^E$. 
% \vspace{-2mm}

\item Remove the SP$^2$ and the SE Attention layers, while just using the proposed CP$^2$ to perform cross-perspective projection based on the coarse pose. There is no further pose refinement. 
% \vspace{-2mm}

\item Replace the SE Attention layers with standard convolution layers to process extracted features. 
% \vspace{-2mm}

\item Remove the pose refinement model SP$^2$, instead using the coarse pose directly as input to the CP$^2$ layer. 
\end{enumerate}
% \vspace{-2mm}

Variant performance is reported in Table \ref{tab:varient} with highlighted the best (bold) and worst (underline) layout estimations. CP$^2$ is the most critical component of our model which makes use of the refined pose from SP$^2$ in order to associate the adjacent view to the anchor view. With CP$^2$ added, we see that both SP$^2$ and SE Attention bring additional substantial gains to our model, with SP$^2$ proving to be slightly more effective than SE Attention.

We further examine the effect of our proposed Mostly Manhattan post processing algorithm. As mentioned in Section~\ref{sec:benchmarks}, due to ZInD's complexity where most visible layouts go beyond the Manhattan world, fully-Manhattan post processing methods such as in HorizonNet \cite{sun2019horizonnet}, and DulaNet \cite{yang2019dula} do not work well. By comparison, the post processing method introduced by AtlantaNet \cite{pintore2020atlantanet} is better suited for general visible layouts. Here, we compare the performance of our proposed Mostly Manhattan post-processing to AtlantaNet's post-processing method, when applied to our network as well as the baselines. The results are reported in Table~\ref{tab:post}. Our Mostly Manhattan method consistently outperforms the AtlantaNet post processing method w/ or w/o GT pose. Also note that even when we apply AtlantaNet post processing to our network's output, we still achieve a better performance compared to the baseline models.

\begin{table}[t]
  \centering \footnotesize
  \caption{  Comparison of different post processing methods.}
    \begin{tabular}{c||c|cc|cc}
    \hline
    \multirowcell{2}{Pose} & \multirowcell{2}{Methods} & \multicolumn{2}{c|}{Most Manhattan PP} & \multicolumn{2}{c}{AtlantaNet PP } \\
\cline{3-6}       &          & 2D IoU & $\delta_i$ & 2D IoU & $\delta_i$ \\ \hline\hline
    \multicolumn{1}{c||}{\multirowcell{5}{w/ \\GT}}  
    & DulaNet~\cite{yang2019dula} &  64.03     & 0.8043      &  62.06   & 0.7899  \\
    & HorizonNet~\cite{sun2019horizonnet}  & 73.35    & 0.8663 & 71.36  & 0.8785  \\
          & HoHoNet~\cite{sun2021hohonet} & 74.25 & 0.8649 & 73.25   & 0.8732    \\
          & LED$^2$Net~\cite{wang2021led2}  & 76.39 & 0.9056 & 75.14 & 0.8849  \\
          & PSMNet (Ours) & \textbf{81.01} & \textbf{0.9238} & \textbf{77.69} & \textbf{0.9159}  \\\hline
    \multicolumn{1}{c||}{\multirowcell{5}{w/o \\GT}} 
    & DulaNet~\cite{yang2019dula} &  59.30     & 0.7828     & 58.90     & 0.7634 \\
    & HorizonNet~\cite{sun2019horizonnet} & 62.79 & 0.8354  & 60.17 & 0.8145 \\
          & HoHoNet~\cite{sun2021hohonet} & 63.31 & 0.8324 & 61.85  & 0.8204 \\
          & LED$^2$Net~\cite{wang2021led2} & 65.81 & 0.8566 & 64.09 & 0.8423 \\
          & PSMNet (Ours) & \textbf{75.77} & \textbf{0.9217} & \textbf{73.22} & \textbf{0.8926}\\\hline
    \end{tabular}%
    \vspace{-2mm}
  \label{tab:post}%
\end{table}%

%------------------------------------------------------------------------

\section{Discussion} \label{sec:discussion}
\begin{figure}[t]
\begin{center}
\includegraphics[width=0.48\textwidth]{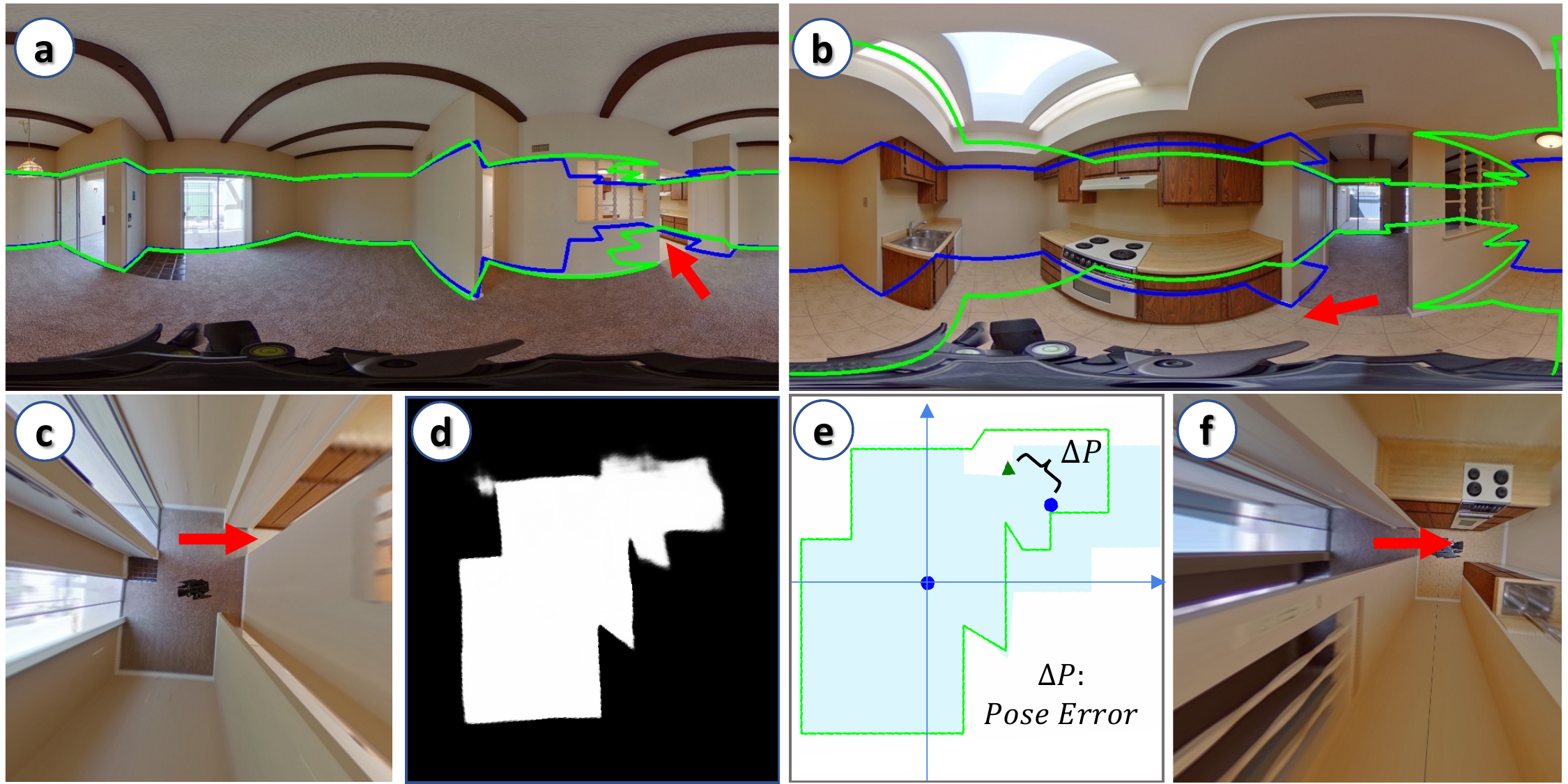} 
% \vspace{-5pt}
\caption{ Illustration of a challenging example, with an overlap score of just 0.27.}
\vspace{-8mm}
\label{fig:failure}
\end{center}
\end{figure}

In Figure \ref{fig:failure}, we share an example which highlights limitations and challenges of our task. Here, the room is not only complex, but there is also a low overlap score (only 0.27) between the two panoramas. As a result, there are few common features between the two views. The {\color{blue}GT} and predicted {\color{green}PSMNet} layouts are shown in Figure \ref{fig:failure} (a) and (b); in (b), boundary misalignment can be seen in panorama 2. We highlight the narrow path connecting the kitchen and living room (with {\color{red}arrows}), which causes low co-visibility. This challenge is further compounded in the perspective views, Figure \ref{fig:failure} (c) and (f), with limited common floor and wall texture between the image pair. 

Figure \ref{fig:failure} (e) displays the error ($\Delta P$) in the refined position of panorama 2. This pose error results in feature misalignment inside our network, which ultimately leads to a noisy predicted segmentation, shown in Figure \ref{fig:failure} (d). Figure \ref{fig:failure} (e) further visualizes the final predicted layout and GT floor segmentation, where we observe a direct correlation between the relative pose error and shifted layout boundary. This particular example also contains a flaw in the data, where the {\color{blue}GT} is missing a portion of floor boundary around the divider between kitchen and living room (which our model recognizes). This means that in actuality the co-visibility and overlap scores are even lower than computed.

%------------------------------------------------------------------------
\section{Conclusion}
We have introduced a novel end-to-end approach for jointly estimating complex room layout from stereo-view panoramas, while refining a noisy relative pose. We adopt a dual-projection backbone architecture to extract features from both equirectangular and perspective-view images. For pose refinement, we propose a transformer-based Stereo Pano Pose (SP$^2$) Network to derive implicit correspondence, and predict refinement parameters by a fully-connected layer. A novel Cross-Perspective Projection (CP$^2$) is crucially designed to project the adjacent panorama view to the anchor view, as well as to align multi-scale equirectangular features for merging in the central segmentation branch. To weight the contribution of features from both views, we apply SE-Attention inspired by \cite{hu2018squeeze}. To evaluate the performance of our method, we introduce baselines based upon currently available state-of-the-art single-perspective layout estimators. Our model demonstrates significant improvements in layout estimation accuracy on a new stereo-view visible layout dataset, derived from ZInD, which will be released to the community.

{\small
\bibliographystyle{ieee_fullname}
\bibliography{egbib}
}

\end{document}